\documentclass{article}

\usepackage[preprint,nonatbib]{nips_2018}

\usepackage[utf8]{inputenc} % allow utf-8 input
\usepackage[T1]{fontenc}    % use 8-bit T1 fonts
\usepackage{amsfonts}       % blackboard math symbols
\usepackage{nicefrac}       % compact symbols for 1/2, etc.
\usepackage{microtype}      % microtypography
\usepackage{graphicx}
\usepackage{subfig}
\usepackage{amsmath}
\usepackage{xfrac}
\usepackage{placeins}
\usepackage{tikz}
\usepackage{wrapfig}
\usepackage{pgfplots}

\usepackage{hyperref}
\hypersetup{hidelinks,breaklinks=true}
\usepackage{breakurl}

\pgfplotsset{compat=1.14}
\DeclareMathOperator{\E}{\mathbb{E}}

\title{A Deep Generative Model for Semi-Supervised Classification with Noisy Labels}

\author{
  Maxime Langevin\thanks{\texttt{maxime.langevin@polytechnique.edu}}, $\,$ Edouard Mehlman\\
  École Polytechnique\\
  \And
  Jeffrey Regier, Romain Lopez,\\\textbf{Michael I. Jordan}, and \textbf{Nir Yosef}\\
  University of California, Berkeley \\
}

\makeatletter
\renewcommand{\@noticestring}{}

\begin{document}
\maketitle
\vspace{-15px}
\section{Introduction}
\vspace{-5px}
Deep generative models perform well at semi-supervised classification when the observed labels are accurate~\cite{kingma2014semi,2017arXiv170609751G}.
Real-world datasets, however, often have imperfectly observed labels, due to mistakes and to genuine ambiguity among classes.
That can severely affect overall classification accuracy.

We propose the Mislabeled VAE (M-VAE), a new deep generative model that explicitly models noisy labels.
While Bayesian approaches to noisy labels have been proposed, none are designed for deep generative models~\cite{NIPS2017_7143,Xiao_2015_CVPR}.
In addition to performing better than existing deep generative models which do not account for label noise,
the derivation of M-VAE gives new theoretical insights into the M1+M2 semi-supervised model~\cite{kingma2014semi}.

\vspace{-2px}
\section{Model}
\vspace{-5px}
For a particular example, let observed random variable $x$ denote the covariates.
Let partially observed random variable $y'$  denote the observed (and possibly incorrect) class label.
Let latent random variable $y$ denote the true class label.
Let $z_1$ and $z_2$ denote additional low-dimensional latent random variables.
Then, the generative process for M-VAE is
\begin{align*}
\*{\*y}, \*z_2 \sim \text{Cat}({\*y})p(\*z_2); \qquad  \*z_1 \sim p_\theta(\*z_1|\*z_2, {\*y});\qquad \*x \sim p_\theta(\*x| \*z_1) \qquad \*y' \sim p_\theta(\*y'| {\*y})
\end{align*} 

Here $p_\theta(\*z_1|\*z_2, {\*y})$ and $p_\theta(\*x| \*z_1)$ are parameterized by deep neural networks. Let $C$ be the number of classes. Then, $p_\theta(\*y'| {\*y})$ is specified by an arbitrary $C \times C$ matrix of conditional probabilities, whose rows sum to one.

\vspace{-2px}
\section{Inference}
\label{inference}
\vspace{-5px}
We use variational inference to approximate the posterior distribution.
We consider approximations that factorize as
\begin{align*}
    q(z_1, z_2, y|x, y') = q(z_1|x,y')q(y|z_1)q(z_2|z_1, y).
\end{align*}
The factors on the right-hand side are all multivariate Gaussian with diagonal covariance matrices.
Their means and covariances are given by deep neural networks.
Then, the evidence lower bound (ELBO) for this model is
\begin{align}
\begin{split}
\mathcal{L}(\phi, \theta) = 
\sum_{n=1}^{N}
&\E_{q_\phi({\*z_1}_n|\*x_n)}[\sum_{k=1}^{C}q_\phi({\*y}_k|{\*z_1}_n)(-\mathrm{KL}(q_\phi({\*z_2}_n|{\*z_1}_n, {\*y}_k) || p(\*z_2)) \\
&+ \E_{q_\phi({\*z_2}_n|{\*z_1}_n, {\*y}_k)}(\log p_\theta({\*z_1}|{\*z_2}_n, {\*y}_k)- \log q_\phi({\*z_1}_n|\*x_n))\\
& + \log p_\theta(\*x_n|{\*z_1}_n) - \mathrm{KL} (q_\phi({\*y}_k|{\*z_1}_n) || p({\*y}_k)] \\
& + \sum_{n=1}^{N_l}\E_{q_\phi({\*z_1}_n|\*x_n)}\sum_{k=1}^{C}q_\phi({\*y}_k|{\*z_1}_n)[q_\phi({\*y}_k|{\*z_1}_n)\log p_\theta(\*y'_n|{\*y}_k)]
\label{terms}
\end{split}
\end{align}
The last term of Equation~\ref{terms} we refer to as the \textit{classification error term} and denote it $\Psi$.

\section{M1+M2 inference is a special case of M-VAE inference}
\label{reinterpret}
\vspace{-6px}
The popular M1+M2 model~\cite{kingma2014semi} is different than the M-VAE graphical model: the former does not model mislabeling.
However, in~\cite{kingma2014semi}, an extra classification penalty term is added to the ELBO for M1+M2 to improve performance,
without any basis in the corresponding graphical model.
By showing an equivalence between the M1+M2 objective function (including the penalty term) and the ELBO for a special case of the M-VAE model,
we show that M1+M2 may be reinterpretted as modeling mislabeling.
This new interpretation gives M1+M2 more basis in probabilistic modeling, which can suggest when M1+M2 will work well.
Additionally, it gives guidance for setting the penalty term's weight $\alpha$ in the M1+M2 objective function,
which previously lacked any link to a probabilistic model.

Consider the following special case of M-VAE:
\vspace{-2px}
\begin{align*}
    p_\theta({\*y'}| {\*y})=
    \begin{cases}
      1-\epsilon, & \text{if}\ {\*y'}={\*y} \\
      \epsilon / (C-1), & \text{otherwise.}
    \end{cases}
\end{align*}
In words, the observed label is correct with probability $1-\epsilon$.
If the observed label is incorrect, then the observed label is selected uniformly from the incorrect classes.

Let $f(\epsilon)  = \log(1-\epsilon) - \log(\sfrac{\epsilon}{(C-1)})$.
For this special case of the M-VAE model, we can rewrite the classification error term as
$\Psi = \E_{q_\phi({\*z_1}_n|\*x_n)}q_\phi(\*y'_n|{\*z_1}_n)f(\epsilon) + \text{Constant}.$
This expression shows that the classification error is weighted by $f(\epsilon)$, a decreasing function of $\epsilon$: the more certainty 
we have on our labels, the more heavily the classification loss is weighted in the ELBO.

Suppose the M1+M2 objective's penalty weight $\alpha = f(\epsilon)$.
Then, the M1+M2 objective function (ELBO + penalty) is equivalent to the M-VAE's ELBO.
Therefore, the M1+M2 inference procedure (though not the M1+M2 graphical model) can be interpreted as modeling label uncertainty,
with corruption probability $\epsilon$ shared by all classes.

\vspace{-3px}
\section{Case study}
\vspace{-8px}
Deep generative models recently have shown promise for interpreting single-cell RNA sequencing (scRNA-seq) data (~\cite{Lopez292037}).
Automatically annotating each cell with its type based on its gene expression profile is an important task.
To our knowledge, deep generative models have not yet been applied to it.
Manually annotating cells is a difficult task for people, so it is common to have datasets with mislabeled cells.

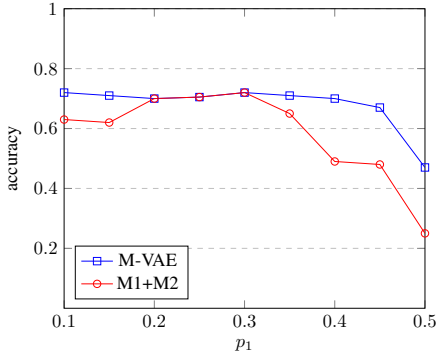
\begin{figure}[h]
\centering
\begin{tikzpicture}[scale=0.7]
\begin{axis}[
    title={},
    xlabel={$p_{1}$},
    ylabel={accuracy},
    xmin=0.1, xmax=0.5,
    ymin=0, ymax=1,
    xtick={0.1, 0.2, 0.3, 0.4, 0.5},
    ytick={.20,.40,.60,.80,1.00},
    legend pos=south west,
    ymajorgrids=true,
    grid style=dashed,
]
\addplot[
    color=blue,
    mark=square,
    ]
    coordinates {
    (0.1, 0.72)(0.15, 0.71)(0.2, 0.7)(0.25, 0.705)(0.3, 0.72)(0.35, 0.71)(0.4, 0.70)(0.45, 0.67)(0.5,0.47)
    };
\addlegendentry{M-VAE}
\addplot[
    color=red,
    mark=o,
    ]
    coordinates {
    (0.1, 0.63)(0.15, 0.62)(0.2, 0.7)(0.25, 0.705)(0.3, 0.72)(0.35, 0.65)(0.4, 0.49)(0.45, 0.48)(0.5,0.25)
    };
\addlegendentry{M1+M2}
\end{axis}
\end{tikzpicture}
\vspace{-5px}
\caption{Test-set classification accuracy. Corruption probability $p_1$ varies and $p_{0} = 0.2$ is fixed.\vspace{-5px}}
\label{theplot}
\end{figure}

We test our model with an scRNA-seq dataset from~\cite{Zeisel1138}.
We consider two cell types: microglia and endothelial.
We corrupt some of the observed labels, mislabeling endothelial with probability $p_{0}$ and microglia with probability $p_{1}$.
Then, we train with those corrupted labels, and predict labels for the whole dataset using ten labeled samples 
per class.\footnote{Our code is available at \burl{https://github.com/maxime1310/fuzzy\_labeling\_scRNA/blob/fuzzy\_labels/CortexNoisyLabels.ipynb}.}

Figure~\ref{theplot} shows that M-VAE outperforms the standard M1+M2 model, which fails to explicitly account for mislabeling, and, even if $\alpha$ is optimal,
fails to account for unequal rates of mislabeling (i.e., $p_0 \ne p_1$).
The advantage of the M-VAE model increases as the imbalance between corruption probabilities $p_0$ and $p_1$ grows.

\FloatBarrier
\bibliographystyle{unsrt}
\bibliography{references}

\begin{thebibliography}{1}

\bibitem{kingma2014semi}
Diederik Kingma, Shakir Mohamed, Danilo~Jimenez Rezende, and Max Welling.
\newblock Semi-supervised learning with deep generative models.
\newblock In {\em Advances in Neural Information Processing Systems}, pages
  3581--3589, 2014.

\bibitem{2017arXiv170609751G}
Jonathan Gordon, José Miguel, and Hernández-Lobato.
\newblock Bayesian semisupervised learning with deep generative models.
\newblock {\em arXiv:1706.09751}, 2017.

\bibitem{NIPS2017_7143}
Arash Vahdat.
\newblock Toward robustness against label noise in training deep discriminative
  neural networks.
\newblock In {\em Advances in Neural Information Processing Systems}, pages
  5596--5605, 2017.

\bibitem{Xiao_2015_CVPR}
Tong Xiao, Tian Xia, Yi~Yang, Chang Huang, and Xiaogang Wang.
\newblock Learning from massive noisy labeled data for image classification.
\newblock In {\em The IEEE Conference on Computer Vision and Pattern
  Recognition}, pages 2691--2699, 2015.

\bibitem{Lopez292037}
Romain Lopez, Jeffrey Regier, Michael Cole, Michael~I. Jordan, and Nir Yosef.
\newblock Bayesian inference for a generative model of transcriptome profiles
  from single-cell {RNA} sequencing.
\newblock {\em bioRxiv}, 2018.

\bibitem{Zeisel1138}
Amit Zeisel, Ana~B. Mu{\~n}oz-Manchado, Simone Codeluppi, Peter L{\"o}nnerberg,
  Gioele La~Manno, Anna Jur{\'e}us, Sueli Marques, Hermany Munguba, Liqun He,
  Christer Betsholtz, Charlotte Rolny, Gon{\c c}alo Castelo-Branco, Jens
  Hjerling-Leffler, and Sten Linnarsson.
\newblock Cell types in the mouse cortex and hippocampus revealed by
  single-cell {RNA}-seq.
\newblock {\em Science}, 347(6226):1138--1142, 2015.

\end{thebibliography}

\end{document}